%% file: main.tex
\documentclass[onefignum,onetabnum]{siamart220329}

\input{ex_shared}

\ifpdf
\hypersetup{
  pdftitle={Title},
  pdfauthor={Yuan Chen}
}
\fi

\begin{document}

\maketitle

\begin{abstract}
We present a new Deep Neural Network (DNN) architecture capable of approximating functions up to machine accuracy. Termed Chebyshev Feature Neural Network (CFNN), the new structure employs Chebyshev functions with learnable frequencies as the first hidden layer, followed by the standard fully connected hidden layers. The learnable frequencies of the Chebyshev layer are initialized with exponential distributions to cover a wide range of frequencies. Combined with a multi-stage training strategy, we demonstrate that this CFNN structure can achieve machine accuracy during training. A comprehensive set of numerical examples for dimensions up to $20$ are provided to demonstrate the effectiveness and scalability of the method.
\end{abstract}



\input{Introduction}
\input{Method}
\input{Examples}
\input{Conclusion}

\bibliographystyle{siamplain}
\bibliography{references}
\end{document}

%% file: ex_shared.tex
\usepackage{lipsum}
\usepackage{amsmath,amsfonts}
\usepackage{graphicx}
\usepackage{epstopdf}
\usepackage{algorithmic}
\usepackage{multirow,graphics,booktabs}
\usepackage{comment}

\ifpdf
  \DeclareGraphicsExtensions{.eps,.pdf,.png,.jpg}
\else
  \DeclareGraphicsExtensions{.eps}
\fi

\definecolor{red2}{RGB}{204,0,0}
\definecolor{blue2}{RGB}{0,103,165}
\hypersetup{colorlinks,linkcolor=[RGB]{0,103,165},citecolor=[RGB]{180,0,0}}
\usepackage[nocompress]{cite}

\newcommand{\RV}[1]{{\color{black}{#1}}}

\newsiamremark{remark}{Remark}
\newsiamremark{hypothesis}{Hypothesis}
\crefname{hypothesis}{Hypothesis}{Hypotheses}
\newsiamthm{claim}{Claim}

\headers{Chebyshev Feature Neural Network}{Zhongshu Xu, Yuan Chen and Dongbin Xiu}

\title{Chebyshev Feature Neural Network for Accurate Function Approximation}

\author{Zhongshu Xu, Yuan Chen and Dongbin Xiu\thanks{E-mail addresses: \texttt{\{xu.4202, chen.11050, xiu.16\}@osu.edu}. Department of Mathematics, The Ohio State University, Columbus, OH 43210, USA. Funding: This work was partially supported by AFOSR FA9550-22-1-0011.}}

\usepackage{amsopn}

\def\be{\begin{equation}}
\def\ee{\end{equation}}

\def\x{\mathbf{x}}

\def\N{\mathbf{N}}

\def\Pi{\mathbf{\Phi}}

\def\fh{\widehat{f}}



%% file: Introduction.tex
\section{Introduction}
In recent years, machine learning has become a prominent research area. It has been widely applied in computer vision \cite{voulodimos2018deep}, data analysis \cite{james2013introduction}, natural language processing \cite{chowdhary2020natural}, recommendation systems \cite{pazzani2007content}. In particular, \RV{D}eep Neural Networks (DNNs) are frequently used in machine learning algorithms. Recently, in the area of scientific computing, DNN-based methods have been widely developed for, for example, solving differential equations \cite{raissi2019physics,yu2018deep,sirignano2018dgm}, inverse problems \cite{jin2017deep,li2019performance, li2020nett}, operator learning \cite{lu2021learning,li2020fourier,brunton2016discovering,qin2019data}, etc. In those problems, high-precision function approximations are an intrinsic requirement for the application of DNNs. 

The mathematical foundation of DNNs rests on the celebrated universal approximation \RV{theorem} \cite{hornik1989multilayer}, which states that DNNs are capable of approximating functions with arbitrary precision. However, it is well known that the practical accuracy of DNNs is lacking for many scientific computing problems. For example, the training accuracy of DNNs can often reach a plateau level of $\mathcal{O}(10^{-5})$ to $\mathcal{O}(10^{-2})$. While this may be adequate for imaging analysis and classification problems, it is insufficient for many scientific computing problems, where high precision is critical for long-term predictions. Efforts have been made to partially understand this, see, for example, the so-called spectral bias phenomenon \cite{rahaman2019spectral,fanaskov2023spectral} or frequency principle \cite{xu2019frequency}, although it is unclear when a complete understanding will be available. Practical algorithms have also been designed to enhance DNNs accuracy, see, for example, Fourier features networks \cite{tancik2020fourier}, multi-stage training \cite{wang2024multi}, Hat activation functions \cite{hong2022activation}. 

In this paper, we present \RV{the} Chebyshev Feature Neural Network (CFNN) architecture and its training procedure to achieve arbitrary training accuracy, down to machine precision should one desire. Our work is inspired by the original work of \cite{wang2024multi}, which employed Fourier feature network and multi-stage training to reach machine precision during training. The contribution of our work lies in the following aspects: (1) The use of Chebyshev features with learnable frequencies in the first hidden layer. This takes advantage of the superior function approximation properties offered by Chebyshev functions and also contains \RV{fewer} hyperparameters than the Fourier features network; (2) a random sampling strategy using exponential distribution for the initialization of the Chebyshev frequency in each of the multi-stage training. This alleviates the difficulty of choosing the frequency parameters in \cite{wang2024multi}; and (3) a systematical study of the method for a large class of functions.  Through extensive numerical study, we demonstrate that CFNN offers excellent approximation capability, for smooth functions or discontinuous functions, in various dimensions.

%% file: Method.tex
\section{Chebyshev Feature Neural Network} \label{Method}

We consider the classical function approximation problem: Let $f: D \mapsto \mathbb{R}$, $D \subset \mathbb{R}^d$, $d \geq 1$, be an unknown function and let $f(\x_1), \dots, f(\x_N)$, $N \geq 1$\RV{, }denote its samples. The objective is to find an approximation function $\fh$ such that $\fh(\x) \approx f(\x),\, \forall x \in D$. A common approach is to seek an \RV{approximation} from a parameterized family of functions $\fh(\cdot;\theta)$ and determine the parameter $\theta$ by minimizing the discrepancy between the samples. The mean squared error (MSE) is often used to quantify this discrepancy:
\be \label{mseloss}
    \min_\theta  \frac{1}{N} \sum_{i=1}^N \left[f(\x_i)-\fh(\x_i\RV{;} \theta)\right]^2. 
\ee

In this work, we focus on using DNN for function approximation. Let $\N:\mathbb{R}^d\mapsto \mathbb{R}$ represent the DNN-based mapping operator. The DNN approximates $f$ by
\be
y = \N(\x; \Theta),
\ee
where $\Theta$ represents all the network's parameters (e.g., weights and biases). The values of $\Theta$ are determined by minimizing the MSE loss in \eqref{mseloss}.

\subsection{Chebyshev Features}

Chebyshev polynomials, a well-known class of orthogonal polynomials, are frequently employed in function approximation, often achieving near-optimal performance; see, for instance, \cite{quarteroni2006numerical, Threfethen2013, Szego1939}. For $n\geq 0$, the $n$th Chebyshev polynomial $T_n$ is defined as
$$
    T_n(x) = \cos(n\arccos(x)), \qquad x\in [-1,1].
$$

In this paper, we generalize Chebyshev polynomials by extending their degree from integers to real numbers. Specifically, we consider:
\be \label{cheby}
   T_\alpha (x) = \cos(\alpha\arccos(x)), \qquad \alpha\in \mathbb{R}^+_0.
\ee
By introducing this extension, we broaden the \RV{``}frequency" domain of Chebyshev polynomials to $\mathbb{R}^+_0$. We refer to these generalized functions as Chebyshev features.

This concept parallels the extension used in the Fourier features network \cite{tancik2020fourier}. However, unlike Fourier features, Chebyshev features do not include a \RV{``}phase" parameter.

\subsection{CFNN Construction}

The Chebyshev Feature Neural Network (CFNN) is designed as a fully connected feedforward network. The dimensionality of the input and output layers corresponds to the dimensions of the target function: the input layer contains $d$ nodes, while the output layer consists of a single node (as in the case of this paper).

\begin{figure}[htbp]
  \centering
  \includegraphics[width=.99\textwidth]{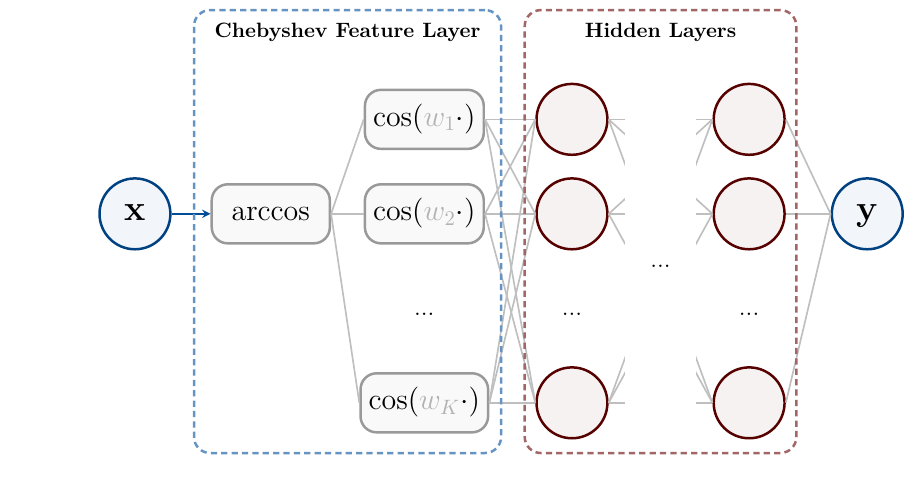}
  \caption{Architechture of the proposed CFNN.}
  \label{fig:CFNN}
\end{figure}

Let $L\geq 1$ represent the number of hidden layers between the input and output layers. For simplicity, we assume each hidden layer contains the same number of $K\geq 1$ nodes.
In CFNN, the first hidden layer employs the Chebyshev features \eqref{cheby} as the activation function, while the remaining hidden layers are the standard feedforward layers. An illustration of the CFNN structure is provided in Figure \ref{fig:CFNN}.
CFNN then defines a mapping operator
\be \label{CFNN_oper}
    \N_{\mathtt{CF}} := \phi_{\mathtt{out}} \circ \phi_{L} \circ \cdots \circ \phi_2 \circ\, \phi_{\mathtt{CF}}.
\ee
Here, $\phi_{\mathtt{CF}}$ is the Chebyshev features operator \eqref{cheby} in the first hidden layer,
\be \label{CF_layer}
    \phi_{\mathtt{CF}}(\x) = \cos(W_{\mathtt{CF}}\arccos(\x)) ,
\ee
where $\x\in \mathbb{R}^d$ is the input, $W_{\mathtt{CF}}=[w_1^T,...,w_K^T]$ the weight matrix, with $K\geq 1$ representing the number of neurons in the next hidden layer. Both $\arccos(\cdot)$ and $\cos(\cdot)$ are applied component-wise.
The remaining layers accomplish the standard operations
$$
\phi_i(\x):= \sigma(W_i\x+b_i), \qquad i=2,\dots,L,
$$
where $\x$ is the outputs from the $(i-1)^{th}$ layer, $W_i$ and $b_i$ are the weights and biases, respectively, and $\sigma(\cdot)$ is the activation function. In this paper, we employ $\tanh(\cdot)$ in these layers.
Finally, the output layer is a linear layer 
$$
\phi_{\mathtt{out}}:=W_{\mathtt{out}}\x+b_{\mathtt{out}}.
$$

\subsection{CFNN Training}

In addition to the network architecture, network training plays a crucial role in ensuring that CFNN achieves machine accuracy. Here, we discuss the key components of CFNN training.

\subsubsection{Multi-stage Training}

The idea of multi-stage training is to train NN on the training \RV{residual} of the NN at the previous stage. It is a general approach that can be adopted for any NN learning. In our setting, let us assume we have trained a CFNN on the given training data set and denote it as $\N_{\mathtt{CF}}^{(0)}(\x)$ at stage 0. 
Let $E_1(\x) = f(\x)-\N_{\mathtt{CF}}^{(0)}(\x)$ be the residue. In the next stage, we train a second CFNN, $\N_{\mathtt{CF}}^{(1)}$, to approximate the scaled residual using the dataset $(\x_i, E_1(\x_i)/\epsilon_1)$, $i=1,..., N$, where the normalizing factor $\epsilon_1$ is defined as
\be
    \epsilon_1=\sqrt{\frac{1}{N} \sum_{i=1}^{N}\left[E_1\left(\x_i\right)\right]^2}.
\ee
The final approximation is then:
\be
    \fh^{(1)}(\x)=\N_{\mathtt{CF}}^{(0)}(\x) + \epsilon_1 \N_{\mathtt{CF}}^{(1)}(\x).
\ee
This iterative process can be readily extended to an arbitrary number of stages $S\geq 1$ to achieve an approximation
\be \label{stage}
    \fh^{(S)}(\x) = \N_{\mathtt{CF}}^{(0)}(\x) + \sum_{s=1}^{S-1}\epsilon_s \N_{\mathtt{CF}}^{(s)}(\x),
\ee
where $\N_{\mathtt{CF}}^{(s)}(\x)$ is trained on the residue data set 
$E_s = f - \fh^{(s-1)}$,
$$
\left(\x_i,\frac{1}{\epsilon_s}E_s(\x_i)\right), \qquad i=1,...,N, \qquad 
$$
with $\epsilon_s$ being normalizing factor of $E_s$.

\subsubsection{Network Initialization}

Network initialization is an important computational aspect.
In \cite{wang2024multi}, scaling parameters were introduced to multi-stage Fourier NN in order to achieve machine accuracy. However, the selection of the scaling parameters depends critically on the properties of the target function. This makes it challenging to apply in practical situations For CFNN, we propose a network initiation approach that can apply to general target functions.

Our initialization method employs random sampling of the network parameters with different distributions.
For the first hidden layer, i.e., the Chebyshev Feature layer \eqref{CF_layer}, the weights $W_{\mathtt{CF}}$ are sampled via exponential distribution, with different parameters at different training stages during the multi-stage training \eqref{stage}. Specifically, at $s^{th}$ stage, we set
$$
W_{\mathtt{CF}}^{(s)} \sim \operatorname{Exp}(\lambda^{(s)}) + c^{(s)}, \qquad s\geq 0.
$$
\RV{In stage 0, $\lambda^{(0)} = 5$ and $c^{(0)} = 0$; in the later stages $\lambda^{(s)} = 5^{1-s}$ and $c^{(s)} = 2 \times 5^{s-1}$}. This initialization encourages the network to first capture the lower-frequency components of the target function, and then in the later stages capture the high-frequency residuals. The exponential distribution increases the likelihood of some of the parameters in the Chebyshev feature layers aligning with the predominant frequencies of the target function and residues.

For the hidden layers after the Chebyshev Feature layer, the weight matrices $W_i$ and biases $b_i$ are initialized with the widely used Xavier method for all the training stages:
$$
W_i \sim \mathcal{N}(0, \sigma^2),\qquad  \sigma = \sqrt{2 / (N_{i-1} + N_i)},
$$
where $N_{i-1}$ and $N_i$ are the numbers of neurons in the previous and next layers. All biases $b_i$ are initialized to zero.

%% file: Examples.tex
\section{Numerical Example} \label{examples}

In this section, we present a comprehensive set of numerical examples to examine the accuracy of the proposed CFNN approach.
In one dimension ($d=1$), we consider the following 6 functions:
\be \label{ex1d}
\begin{split}
    &  f_1(x)=x;  \\
    &  f_2(x)= \sin(2x+1)+0.2e^{1.3x}; \\\
    &  f_3(x)= \left|\sin(\pi x)\right|^2; \\
    &   f_4(x)=\left(1-\frac{x^2}{2}\right) \cos \left[m\left(x+0.5 x^3\right)\right], \qquad m=30;  \\
    &  f_5(x)= \left|x\right|; \\
    &  f_6(x)= \text{sign}(x).
    \end{split}
\ee
In multi-dimension ($d>1$), we test the following 3 functions that are widely used for multi-dimensional function approximation:
\be \label{exnd}
\begin{split}
  &  f_7(\mathbf{x})=\sum_{i=1}^d x_i; \\
    & f_8(\mathbf{x})=\exp \left(-\sum_{i=1}^d \sigma_i^2\left(\frac{x_i+1}{2}-\omega_i\right)^2\right);  \\
    &  f_9(\mathbf{x})=\sum_{i=1}^N \alpha_i\exp \left(-\sum_{j=1}^d \sigma_{ij}^2\left(\frac{x_j+1}{2}-\omega_{ij}\right)^2\right).  
\end{split}
\ee
\RV{The domain of these functions will be specified in the following subsections.} The parameters $\alpha$, $\sigma$, and $\omega$ determine the behavior of the functions and are specified in each example. For the Gaussian peak function $f_8$, we set $\sigma_i = 1$ and $\omega_i=1$, $\forall i$. For the multi-modal function $f_9$, we set $N=10$, $\sigma_{ij}\equiv 1$, and randomly sample $\omega_{ij}$ uniformly in $[-1,1]$ and $\alpha_i$ uniformly in $[-10,10]$.

In most examples, we employ CFNN with 3 hidden layers, each containing 40 neurons (including the first Chebyshev Feature Layer). For the network training at each stage,
we employ the Adam optimizer for \RV{5,000} epochs, followed by the L-BFGS optimizer for an additional \RV{20,000} epochs. The Adam optimization utilizes an exponential decay learning rate, which is initialized at 0.01 and decays by a factor of 0.97 for every 100 epochs. \RV{We conducted tests on both single and double precision, and similar results were observed. In this paper, we only report double-precision results.}

\subsection{One-dimension Examples}
\RV{For one-dimensional examples, we set the domain as $D=[-1,1]$. The training set consists of 3,000 equidistant points in $D$. To evaluate the generalization capability, we test the trained network on a separate set of 10,000 equidistant points in $D$.}

\subsubsection{Smooth Functions}

In this section, we consider the four smooth functions  $f_1$, $f_2$, $f_3$, and $f_4$ in \eqref{ex1d}.

\begin{figure}[htbp]
  \centering
  \includegraphics[width=\textwidth]{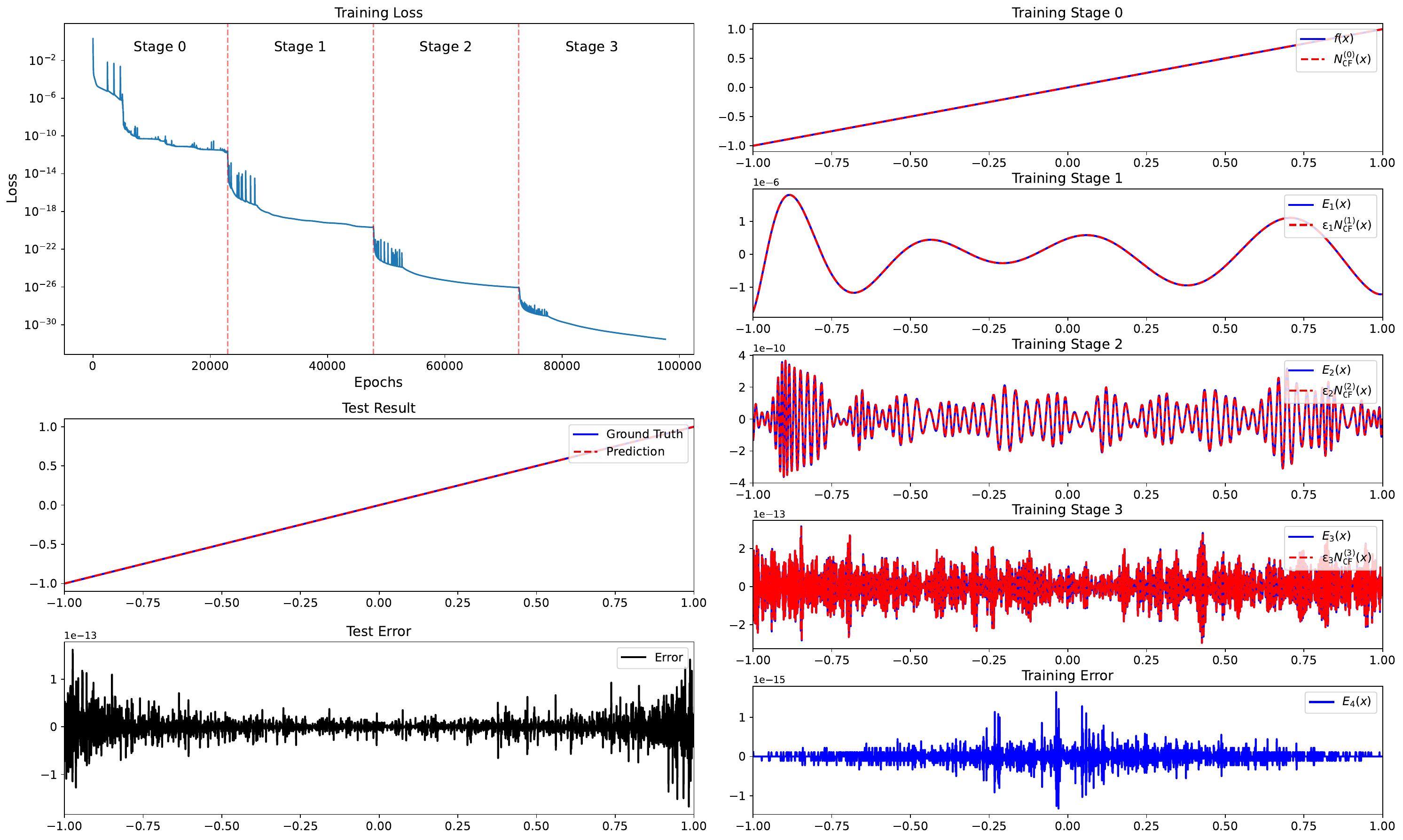}
  \caption{Results of $f_1(x)$. \RV{Left top: training loss history across all stages; Left bottom: test data ground truth vs prediction and prediction error; Right: training results at four stages showing the network output (red dashed) against the ground truth/residual (blue solid) at each stage.}}
  \label{fig: f1}
\end{figure}

For the linear function $f_1(x)$,  we conduct its approximation with a 4-stage CFNN. \RV{The combined results are presented in Figure \ref{fig: f1}. On the left panel of the figure, the training loss is displayed at the top, where we observe that the training RMSE is reduced to $\mathcal{O}(10^{-30})$ through 4 stages of training. The target function and the CFNN approximation are plotted in the middle, where no visible difference can be observed. The point-wise difference, termed test error, is shown in the bottom plot of the left panel. We observe the point-wise errors are of $\mathcal{O}(10^{-13})$.

On the right panel of Figure \ref{fig: f1}, we present the approximation results on the \emph{training data set} for all 4 stages of CFNN, from top to bottom. Note that at training stage 0, the target function is the original function, whereas at later stages the target functions are the residues of the CFNN approximations of the previous stages. The point-wise error is shown at the very bottom. We observe that for such an extremely oscillatory target function of Training Stage 3, the CFNN can achieve approximation error of $\mathcal{O}(10^{-15})$, the machine epsilon level for double precision computation.
}

\begin{figure}[htbp]
  \centering
  \includegraphics[width=\textwidth]{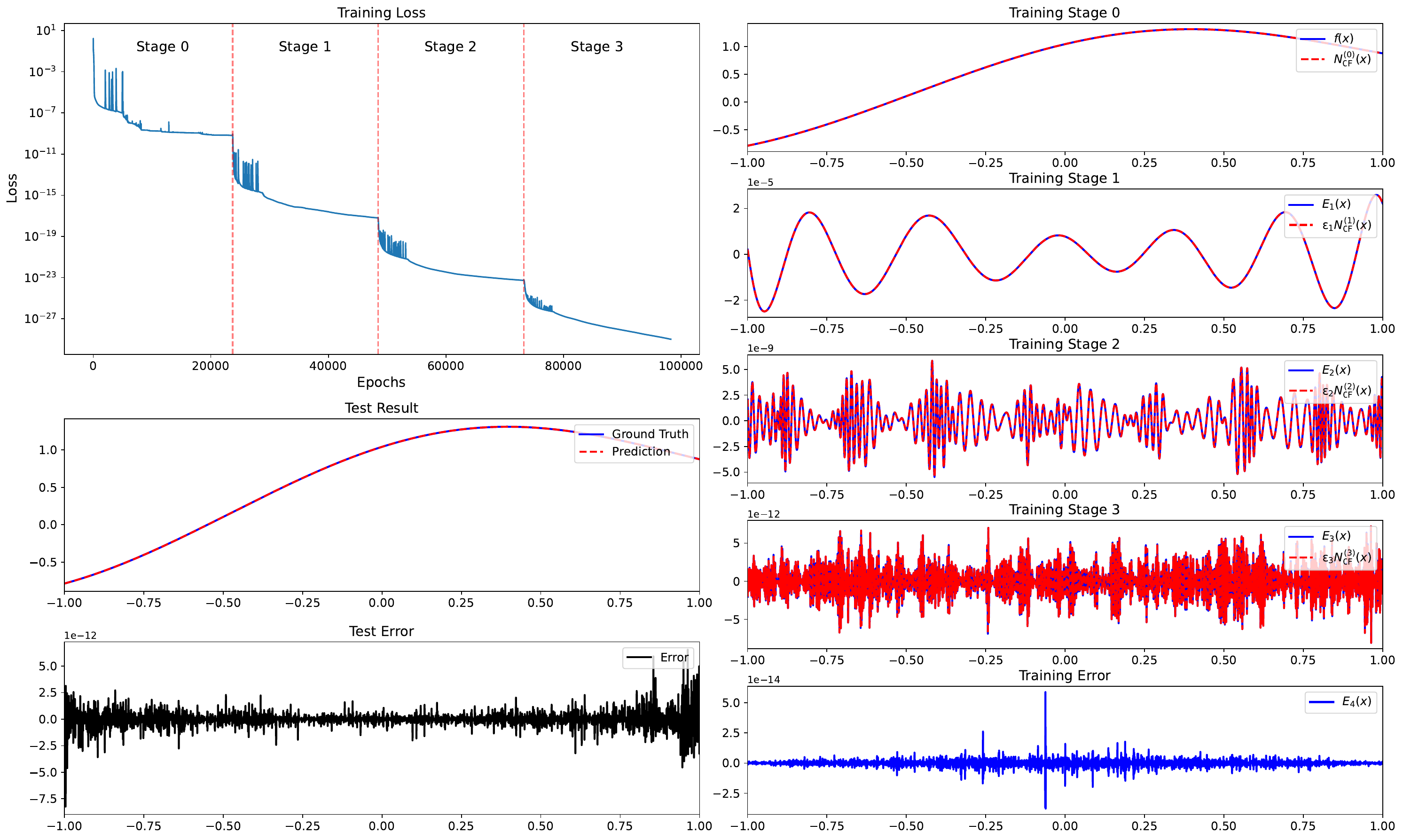}
  \caption{Results of $f_2(x)$. \RV{Left top: training loss history across all stages; Left bottom: test data ground truth vs prediction and prediction error; Right: training results at four stages showing the network output (red dashed) against the ground truth/residual (blue solid) at each stage.}}
  \label{fig: f2}
\end{figure}

We now consider the smooth nonlinear function $f_2$, whose results are shown in Figure \ref{fig: f2}. \RV{The left panel shows (from top to bottom) the training loss decay over the 4 training stages; the CFNN function approximation at the testing set (along with the target function $f_2$); and point-wise error of the CFNN approximation. We observe that the training loss decays to $\mathcal{O}(10^{-28})$ and trained CFNN approximation errors is at $\mathcal{O}(10^{-12})$. The right panel shows the CFNN approximation at the training data set at each training stage (from top down), along with the point-wise error of the last stage CFNN approximation (at the bottow) which is at $\mathcal{O}(10^{-14})$.}

\begin{figure}[htbp]
  \centering
  \includegraphics[width=\textwidth]{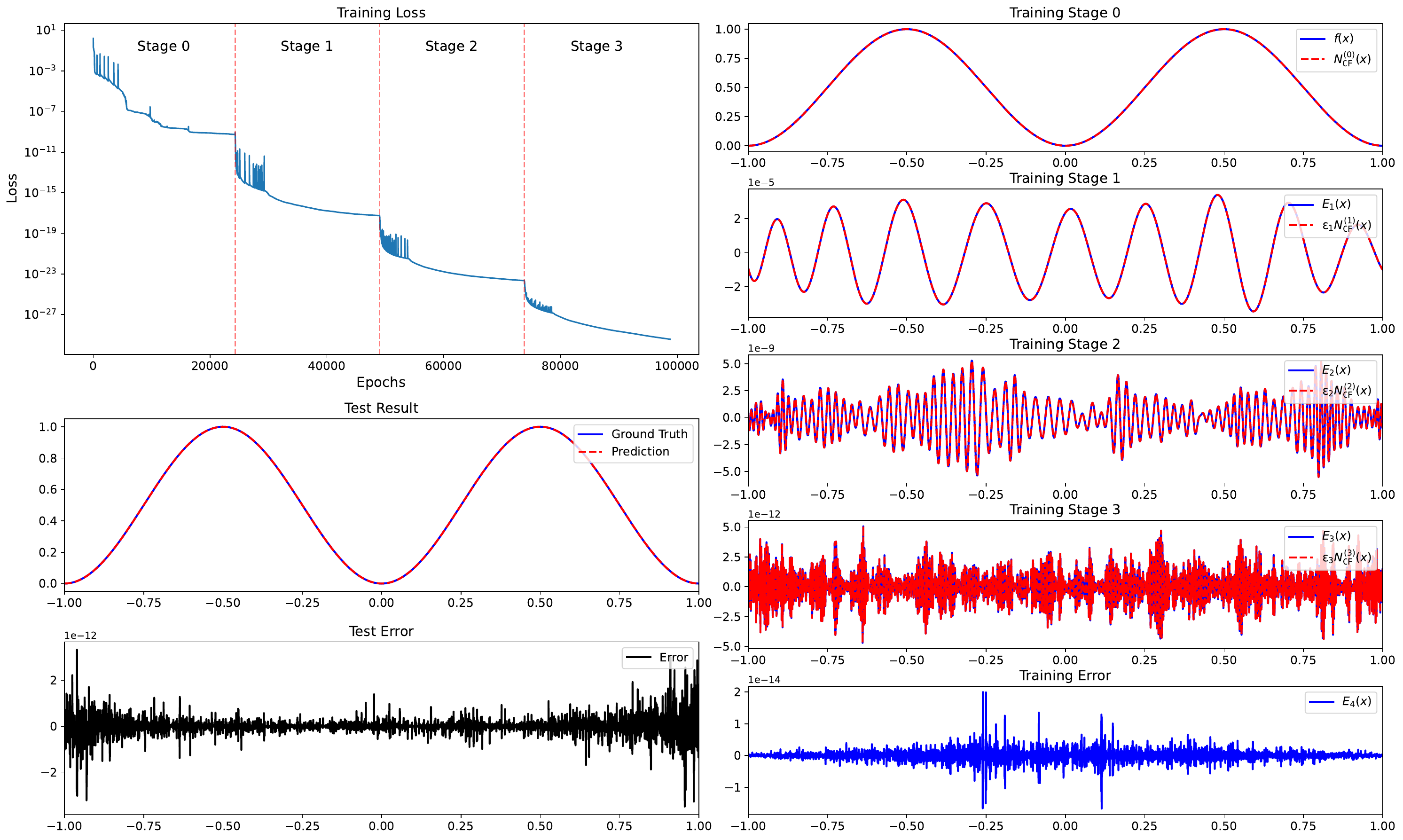}
  \caption{Results of $f_3(x)$. \RV{Left top: training loss history across all stages; Left bottom: test data ground truth vs prediction and prediction error; Right: training results at four stages showing the network output (red dashed) against the ground truth/residual (blue solid) at each stage.}}
  \label{fig: f3}
\end{figure}

\begin{figure}[htbp]
  \centering
  \includegraphics[width=\textwidth]{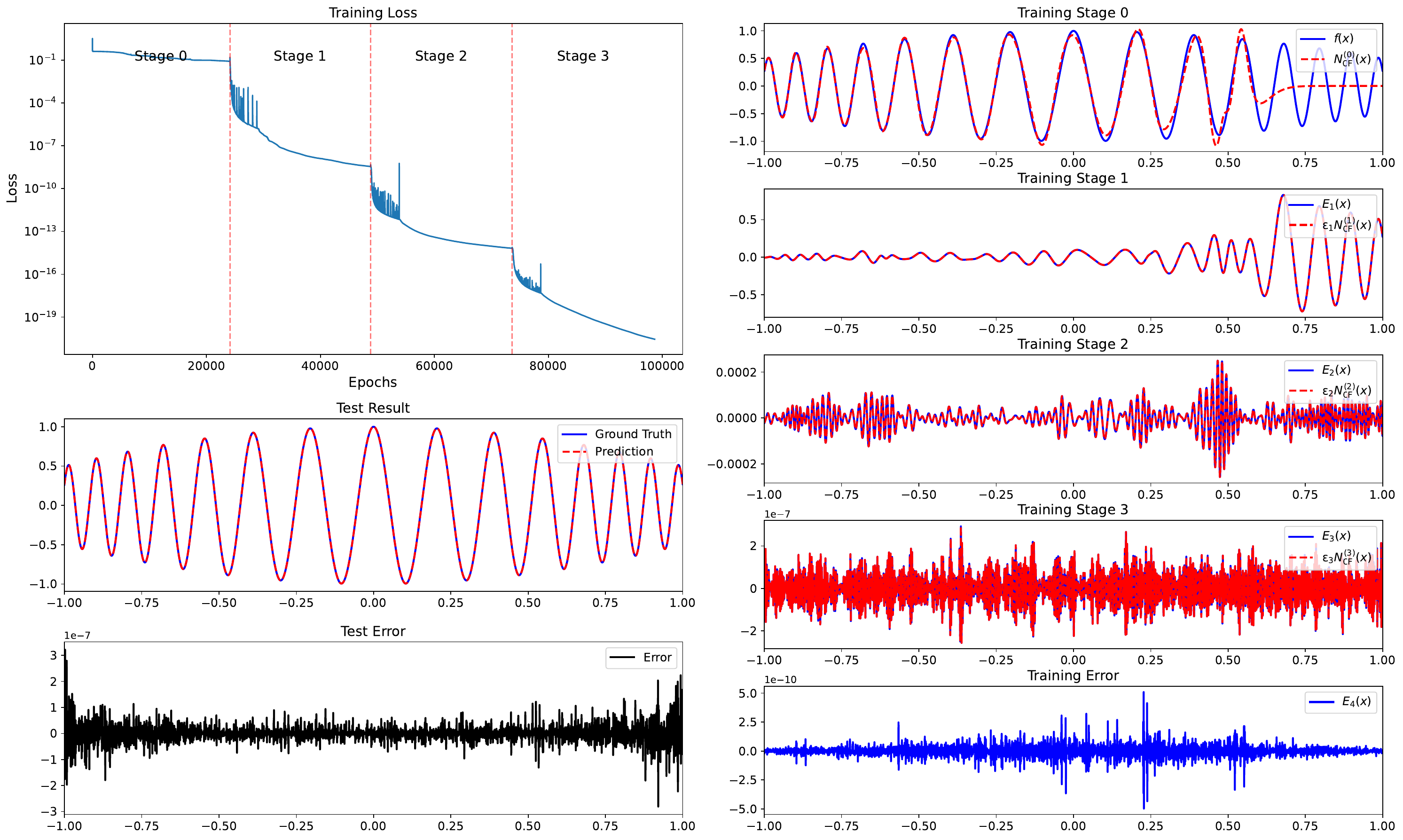}
  \caption{Results of $f_4(x)$. \RV{Left top: training loss history across all stages; Left bottom: test data ground truth vs prediction and prediction error; Right: training results at four stages showing the network output (red dashed) against the ground truth/residual (blue solid) at each stage.}}
  \label{fig: f4}
\end{figure}

The numerical results for $f_3$ are shown in Figure \ref{fig: f3}, and $f_4$ in Figure \ref{fig: f4}. In each plot, we show in the left panel the training loss decay, the trained CFNN approximation, along with its point-wise errors; in the right panel, we show the CFNN approximation after each stage, along with the pointwise error of the last stage.

While the results of $f_3$ are similar to those of $f_1$ and $f_2$, it is worth noting the results for $f_4$, which is a fairly oscillatory function (with $m=30$).  We observe that the one-stage approximation (at Training stage 0) is not very accurate, as shown on the top of the right panel of Figure \ref{fig: f4}. 
The DNN approximation missed a few wavefronts completely. We remark that this is quite common for the standard DNN function approximation.
With the multi-stage learning, the DNN approximation becomes increasingly accurate. \RV{After 4 stages of training,} the pointwise approximation error is at $\mathcal{O}(10^{-10})$ over the training data set. (See the bottom of the right panel of Figure \ref{fig: f4}.) The pointwise approximation error over the testing data set is of $\mathcal{O}(10^{-7})$. (See the bottom of the left panel of Figure \eqref{fig: f4}.) This error is higher than the errors in the previous examples. However, it is not unexpected because of the oscillatory nature of the function.  

\RV{\subsubsection{Ablation Study}

To further examine the proposed CFNN structure, we perform an ablation study to the relatively more challenging function $f_4$. The objective is to understand the roles of the components in the CFNN structure. In particular, we consider the following cases:
\begin{enumerate}
    \item[Case 1:] In the first layer of CFNN, change its activation function from the Chebyshev function \eqref{CF_layer} to the standard $\tanh(\cdot)$ activation function, which is initialized by the standard  Xavier initialization. By doing so, the network becomes a standard feedforward DNN with multi-stage training. This case shall examine the effect of the Chebyshev function used in the first layer of CFNN.
    \item[Case 2:] Since CFNN at each stage does not share parameters, one may consider the multi-stage CFNN a much larger DNN. To examine this, we employ a CFNN with 160 nodes per layer, so that this CFNN has roughly the same number of hyperparameters as the one used in Figure \ref{fig: f4}. We also conduct a standard one-stage training for up to 10,000 epochs. (The Chebyshev Feature layer is initialized with $W_{\mathtt{CF}} \sim \operatorname{Exp}(5^{-3})$.) This case shall examine the impact of multi-stage training in CFNN.
\end{enumerate}

The results of the two ablation cases are shown in Figure \ref{fig: ablation}, where Case 1 is on the left and Case 2 on the right. Compared with the original 4-stage CFNN (Figure \ref{fig: f4}), these two cases demonstrate notable performance degradation. In particular, Case 1 can only achieve pointwise approximation error of $\mathcal{O}(10^{-2})$, whereas Case 2 of $\mathcal{O}(10^{-5})$. Compared to the $\mathcal{O}(10^{-10})$ error achieved by the proposed multi-stage CFNN, we conclude that both the Chebyshev function layer and the multi-stage training are essential in achieving high accuracy, with the Cheyshev function layer playing perhaps a relatively larger role.

\begin{figure}[htbp]
  \centering
  \includegraphics[width=\textwidth]{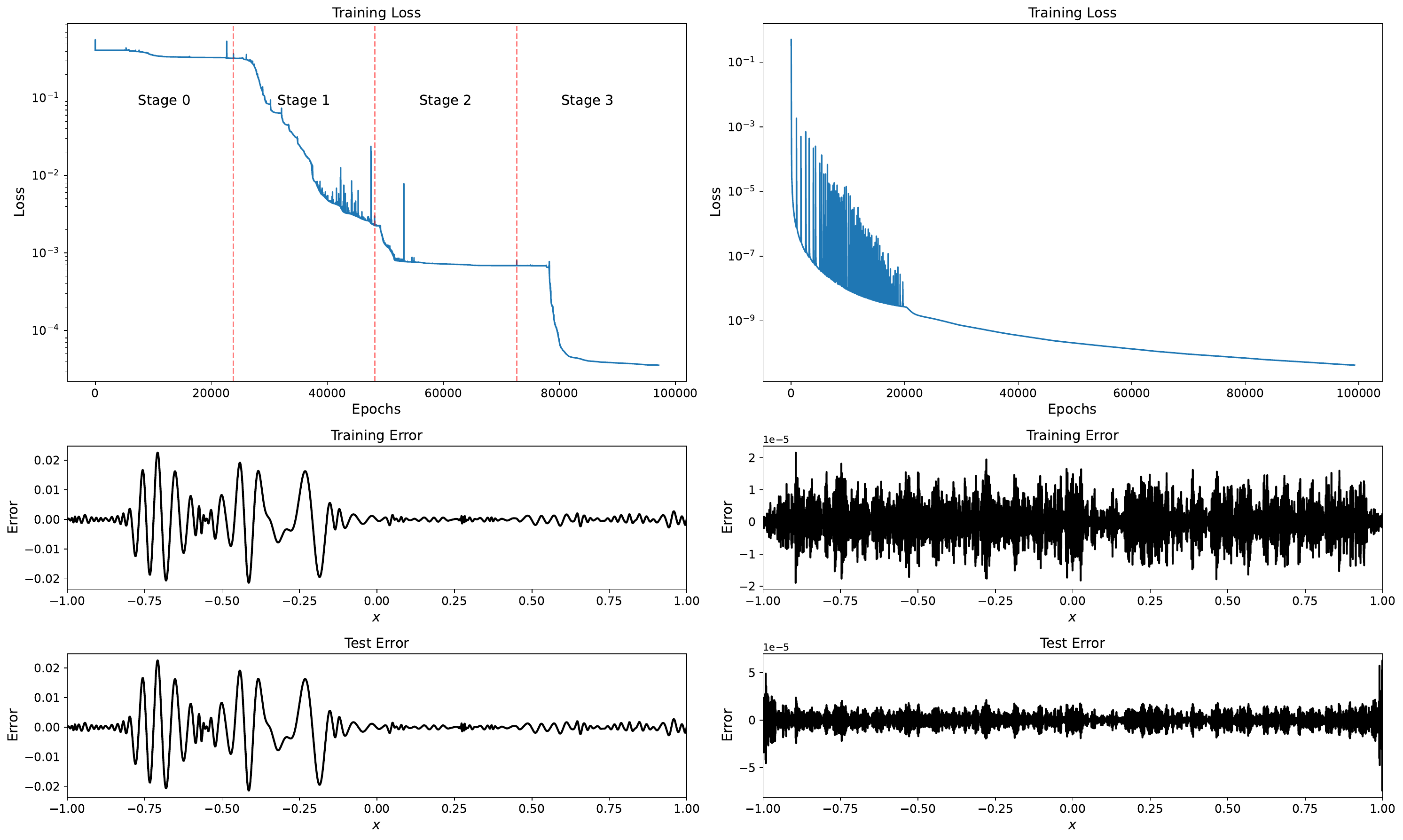}
  \caption{\RV{Results of ablation tests. Left: Case 1; Right: Case 2. From top to bottom, training error decay; Pointwise approximation error at the training data; Pointwise error at the testing data.}}
  \label{fig: ablation}
\end{figure}}

\subsubsection{Nonsmooth Functions}

\begin{figure}[htbp]
  \centering
  \includegraphics[width=\textwidth]{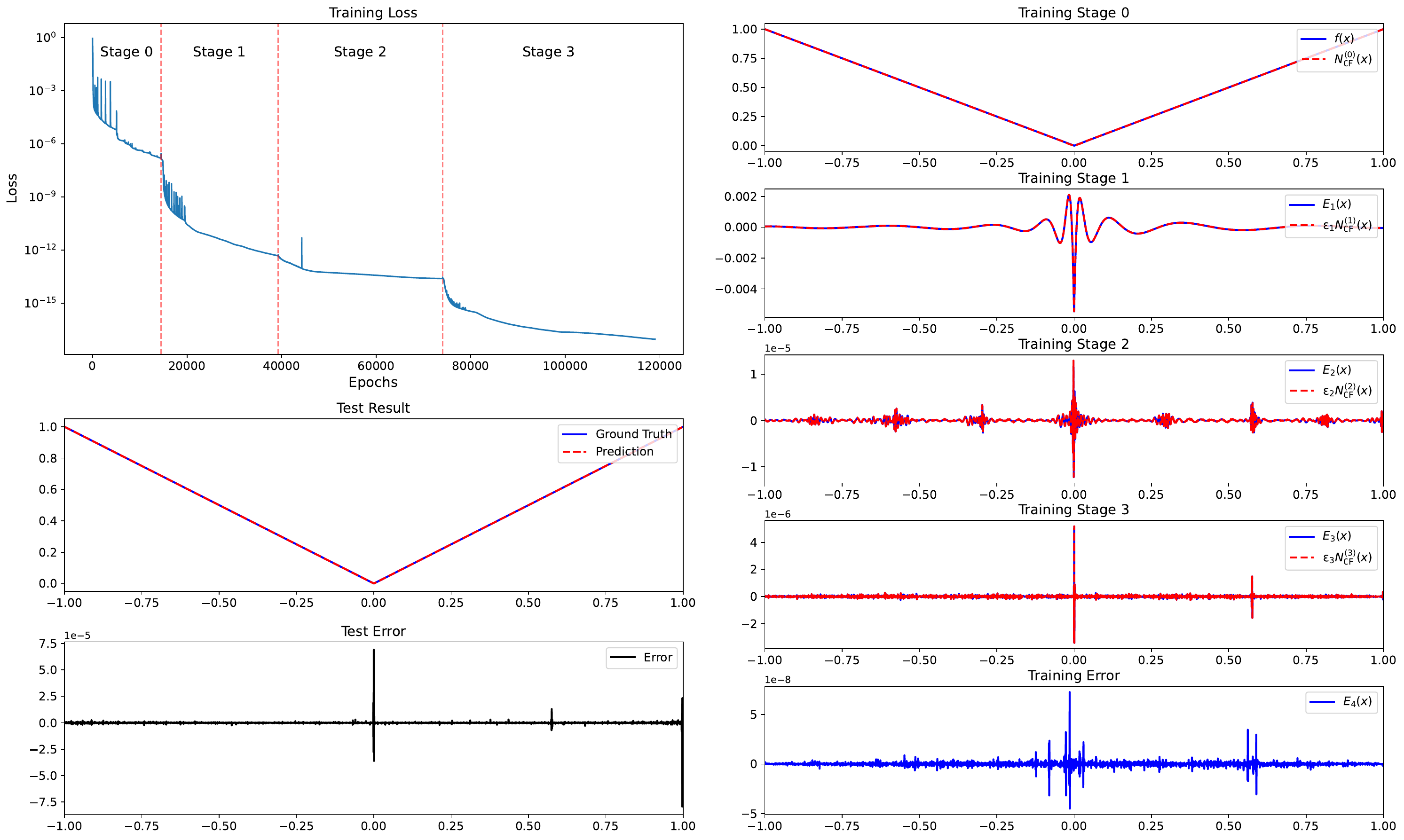}
  \caption{Results of $f_5(x)$. \RV{Left top: training loss history across all stages; Left bottom: test data ground truth vs prediction and prediction error; Right: training results at four stages showing the network output (red dashed) against the ground truth/residual (blue solid) at each stage.}}
  \label{fig: f5}
\end{figure}

\begin{figure}[htbp]
  \centering
  \includegraphics[width=\textwidth]{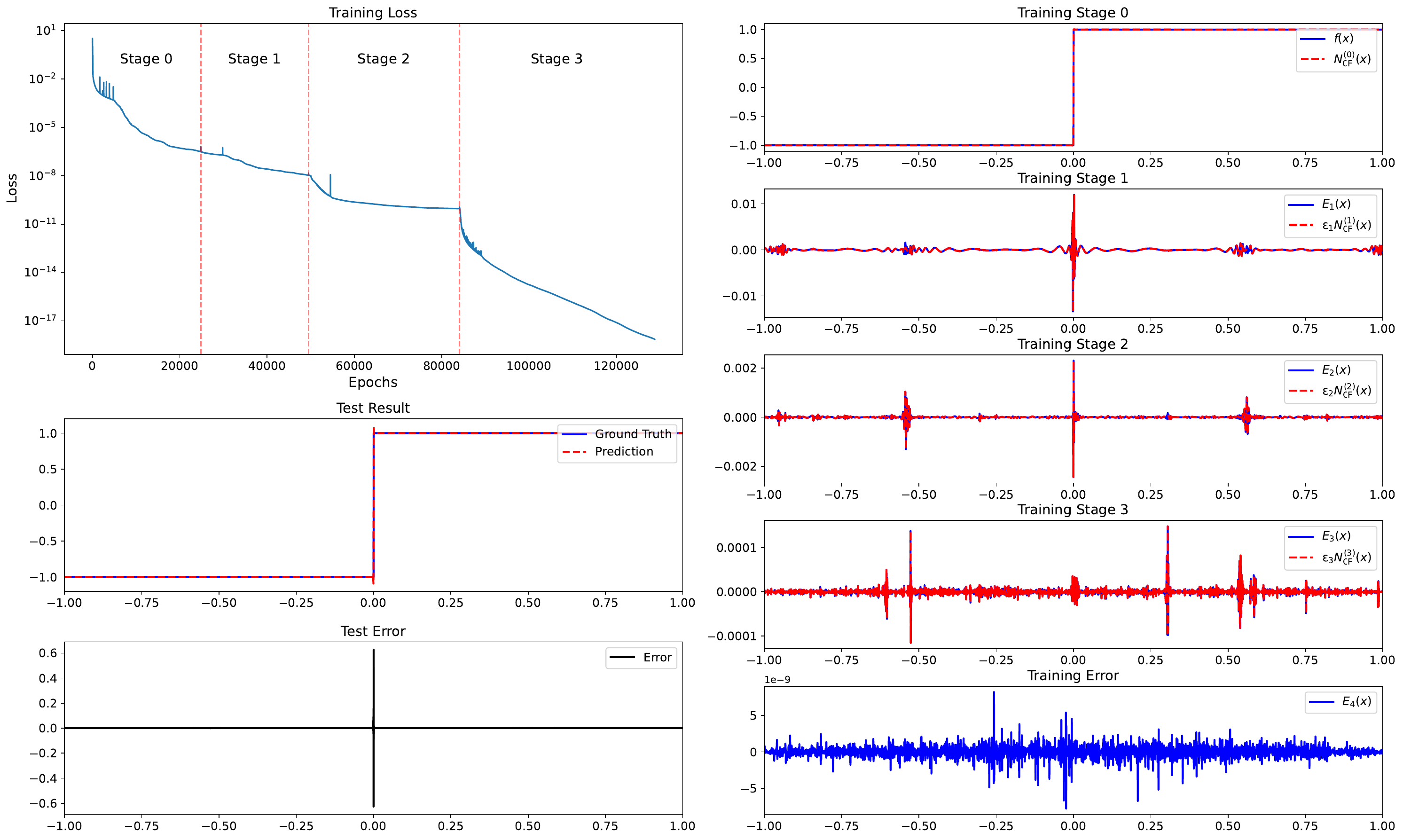}
  \caption{Results of $f_6(x)$. \RV{Left top: training loss history across all stages; Left bottom: test data ground truth vs prediction and prediction error; Right: training results at four stages showing the network output (red dashed) against the ground truth/residual (blue solid) at each stage.}}
  \label{fig: f6}
\end{figure}

We now consider the two non-smooth functions in \eqref{ex1d}, the $C^0$ function $f_5$ and the discontinuous function $f_6$. Both are approximated by 4-stage CFNN. The results for $f_5$ are shown in Figure \ref{fig: f5}, and the results for $f_6$ in Figure \ref{fig: f6}. For $f_5$, \RV{the training loss decays to $\mathcal{O}(10^{-17})$ and the training error is at $\mathcal{O}(10^{-8})$. For the discontinuous function $f_6$, the training loss decays to $\mathcal{O}(10^{-18})$ and the training error is at $\mathcal{O}(10^{-9})$.}
These errors are larger than the machine accuracy errors in the previous examples for smooth functions. This is expected as both functions are non-smooth at $x=0$.
On the other hand, achieving approximation error at \RV{$\mathcal{O}(10^{-8})$} for discontinuous functions is quite remarkable, as most function approximation methods can not achieve this level of accuracy without special treatment of the singularity.

\subsection{Multi-dimension Examples}

For high-dimensional examples, we consider the functions in \eqref{exnd} in $[-1,1]^d$ with various values of $d>1$. We use the proposed CFNN with 4 hidden layers and 40 neurons per layer (including the first Chebyshev Feature Layer). The CFNN training is conducted over a set of $20,000$ uniformly distributed random points. For validation, the approximation errors are computed over an independent set of $10,000$ uniformly distributed random points. \RV{Since it is difficult to visualize high dimensional functions, hereafter we will only report RMSE (root mean squared error) in each case, where ``training error" refers to RMSE over the training data points and ``testing error" refers to RMSE over the testing data points.}

\begin{figure}[htbp]
  \centering
  \label{fig:errorstageEx1}
  \includegraphics[width=.48\textwidth]{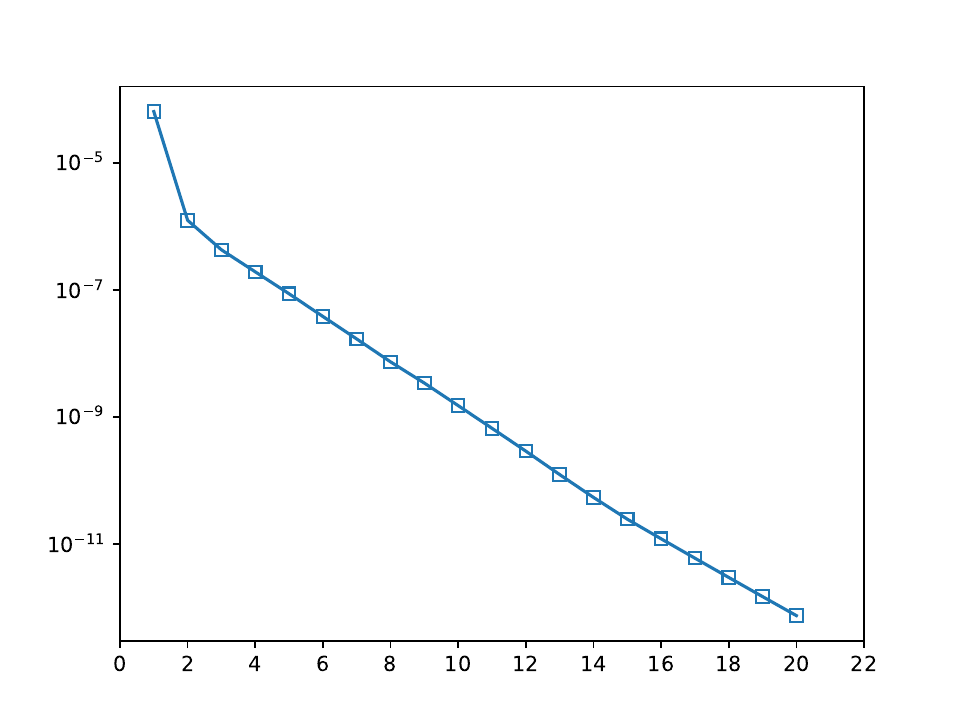}
  \includegraphics[width=.48\textwidth]{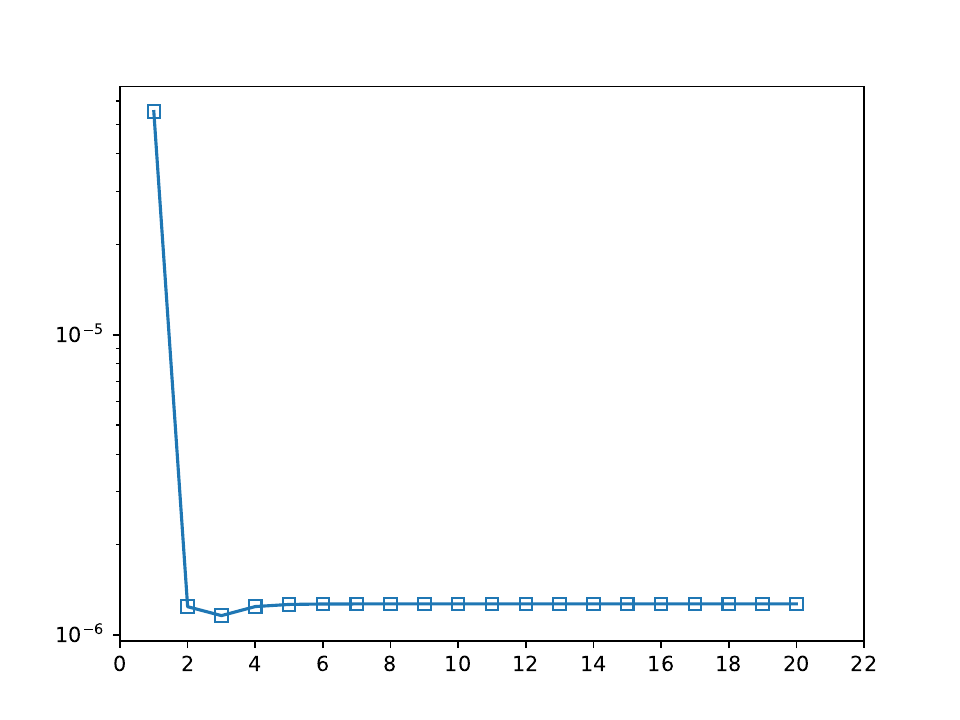}
  \includegraphics[width=.48\textwidth]{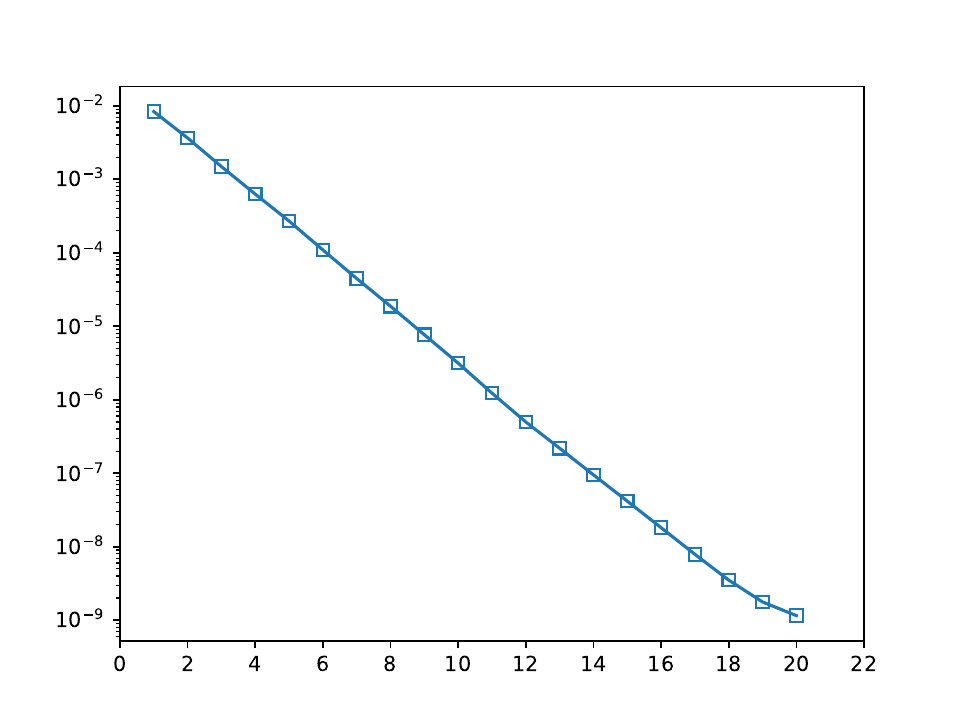}
  \includegraphics[width=.48\textwidth]{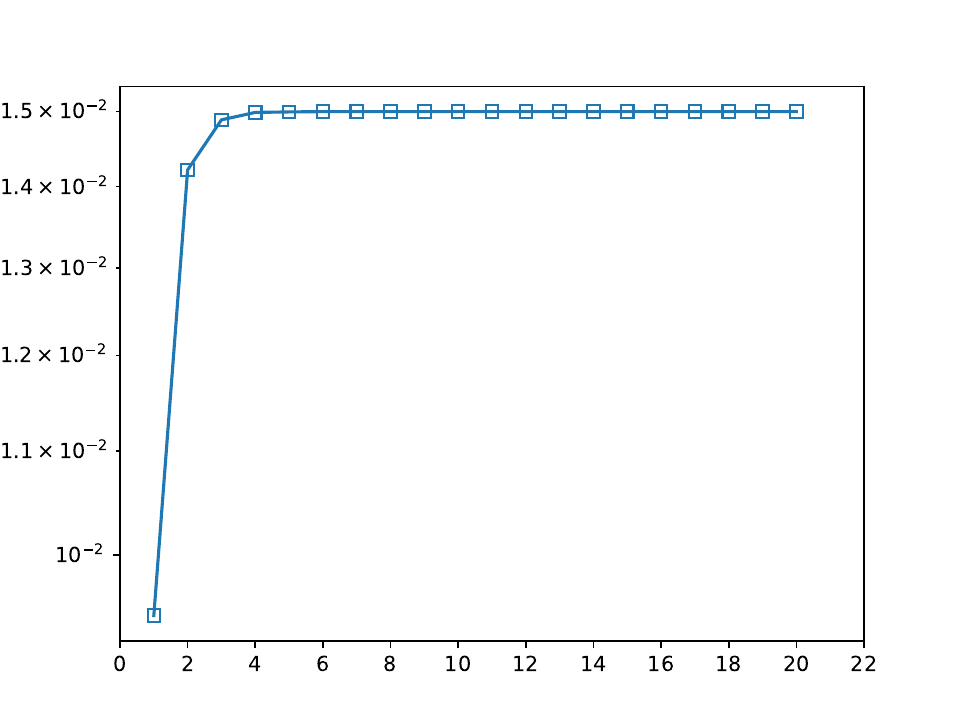}
  \caption{The relative training and testing errors {\em vs.} stages, for $f_7$ in \eqref{exnd} Upper left: training error for $d=2$; Upper right: testing error for $d=2$; Lower left: training error for $d=20$; Lower right: testing error for $d=20$.}
\end{figure}

Numerical tests are performed in dimensions $d=2, 5, 10, 20$, and with learning stages as high as $s=20$.
In Figure \ref{fig:errorstageEx1}, we show both the relative training errors and relative validation errors at each stage for $f_7$ in \eqref{exnd}, at dimension $d=2$ and $d=20$. We observe exponential decay of the training errors with respect to the increase of training stages. The validation errors quickly saturate. This is because the validation errors are computed on a set of fixed, albeit randomly generated, points, whose geometrical property determines the achievable approximation accuracy.

Very similar approximation behaviors are observed for all the 3 examples in \eqref{exnd}. Therefore, we opt not to display all the plots. Instead, we tabulate the relative training errors and relative validation errors for $f_7$ in Table \ref{tabmlinear}, $f_8$ in Table \ref{tabgaussian}, and $f_9$ in Table \ref{tabmgaussian}.
\begin{table}[htbp]
\caption{Summary of Errors for $f_7$ in \eqref{exnd}}
\label{tabmlinear}
\resizebox{\textwidth}{!}{
\begin{tabular}{cccccccc}
\toprule
$d$        & \textbf{Error} & \textbf{Stage 1} & \textbf{Stage 4} & \textbf{Stage 8} & \textbf{Stage 12} & \textbf{Stage 16} & \textbf{Stage 20} \\ \hline
\multirow{2}{*}{2}  & Training & 6.3941E-05       & 1.9261E-07       & 7.3827E-09       & 2.9078E-10      & 1.2144E-11       & 7.5117E-13       \\ \cline{2-8} 
                    & Testing  & 5.5311E-05       & 1.2439E-06       & 1.2699E-06       & 1.2698E-06       & 1.2698E-06       & 1.2698E-06       \\ \hline
\multirow{2}{*}{5}  & Training & 3.9762E-04       & 2.0109E-05       & 7.8913E-07       & 3.0380E-08       & 1.4001E-09       & 7.6904E-11       \\ \cline{2-8} 
                    & Testing  & 3.3952E-04       & 2.6595E-04       & 2.6726E-04       & 2.6724E-04       & 2.6724E-04       & 2.6724E-04       \\ \hline
\multirow{2}{*}{10} & Training & 2.5685E-03       & 2.3697E-04       & 8.3021E-06       & 2.9867E-07       & 1.3257E-08       & 7.6304E-10       \\ \cline{2-8} 
                    & Testing  & 2.4612E-03       & 3.9598E-03       & 3.9698E-03       & 3.9698E-03       & 3.9698E-03       & 3.9698E-03       \\ \hline
\multirow{2}{*}{20} & Training & 8.3437E-03       & 6.2842E-04       & 1.8745E-05       & 4.9859E-07       & 1.8172E-08       & 1.1561E-09       \\ \cline{2-8} 
                    & Testing  & 9.4638E-03       & 1.4985E-02       & 1.4996E-02       & 1.4996E-02       & 1.4996E-02       & 1.4996E-02       \\ \bottomrule
\end{tabular}}
\end{table}

\begin{table}[htbp]
\caption{Summary of Errors for $f_8$ in \eqref{exnd}}
\label{tabgaussian}
\resizebox{\textwidth}{!}{
\begin{tabular}{cccccccc}
\toprule
$d$        & \textbf{Error} & \textbf{Stage 1} & \textbf{Stage 4} & \textbf{Stage 8} & \textbf{Stage 12} & \textbf{Stage 16} & \textbf{Stage 20} \\ \hline
\multirow{2}{*}{2}  & Training & 1.2234E-04       & 6.1596E-07       & 2.3968E-08       & 8.5897E-10       & 4.0117E-11       & 2.4655E-12       \\ \cline{2-8} 
                    & Testing  & 1.0396E-04       & 4.4164E-06       & 4.4850E-06       & 4.4853E-06       & 4.4853E-06       & 4.4853E-06       \\ \hline
\multirow{2}{*}{5}  & Training & 9.9211E-04       & 8.3535E-05       & 3.2130E-06       & 1.1872E-07       & 5.7236E-09       & 3.2697E-10       \\ \cline{2-8} 
                    & Testing  & 8.7179E-04       & 1.1540E-03       & 1.1574E-03       & 1.1574E-03       & 1.1574E-03       & 1.1574E-03       \\ \hline
\multirow{2}{*}{10} & Training & 3.6173E-03       & 3.1649E-04       & 1.1602E-05       & 4.3113E-07       & 1.9065E-08       & 1.0722E-09       \\ \cline{2-8} 
                    & Testing  & 7.4386E-03       & 8.8001E-03       & 8.8122E-03       & 8.8124E-03       & 8.8124E-03       & 8.8124E-03       \\ \hline
\multirow{2}{*}{20} & Training & 2.3357E-02       & 1.9097E-03       & 6.0061E-05       & 1.7198E-06       & 5.9873E-08       & 3.8842E-09      \\ \cline{2-8} 
                    & Testing  & 1.3465E-01       & 1.3750E-01       & 1.3754E-01       & 1.3754E-01       & 1.3754E-01       & 1.3754E-01       \\ \bottomrule
\end{tabular}}
\end{table}

\begin{table}[htbp]
\caption{Summary of Errors for $f_9$ in \eqref{exnd}}
\label{tabmgaussian}
\resizebox{\textwidth}{!}{
\begin{tabular}{cccccccc}
\toprule
$d$        & \textbf{Error} & \textbf{Stage 1} & \textbf{Stage 4} & \textbf{Stage 8} & \textbf{Stage 12} & \textbf{Stage 16} & \textbf{Stage 20} \\ \hline
\multirow{2}{*}{2}  & Training & 1.3982E-04       & 8.4385E-07       & 3.2461E-08       & 1.2081E-09       & 5.6564E-11       & 3.5100E-12       \\ \cline{2-8} 
                    & Testing  & 1.1950E-04       & 5.8842E-06       & 5.9729E-06       & 5.9728E-06       & 5.9728E-06       & 5.9728E-06       \\ \hline
\multirow{2}{*}{5}  & Training & 3.1952E-03       & 3.5137E-04       & 1.2766E-05       & 4.5370E-07       & 2.1068E-08       & 1.1673E-09       \\ \cline{2-8} 
                    & Testing  & 3.1045E-03       & 4.4888E-03       & 4.5006E-03       & 4.5005E-03       & 4.5006E-03       & 4.5006E-03       \\ \hline
\multirow{2}{*}{10} & Training & 2.6111E-02       & 2.3253E-03       & 8.4740E-05       & 2.9449E-06       & 1.2474E-07       & 7.0201E-09       \\ \cline{2-8} 
                    & Testing  & 4.0942E-02       & 5.1587E-02       & 5.1669E-02       & 5.1670E-02       & 5.1670E-02       & 5.1670E-02       \\ \hline
\multirow{2}{*}{20} & Training & 7.2815E-02       & 5.6877E-03       & 1.6364E-04       & 4.5827E-06       & 1.6399E-07       & 1.0466E-08       \\ \cline{2-8} 
                    & Testing  & 6.0428E-01       & 6.1028E-01       & 6.1019E-01       & 6.1019E-01       & 6.1019E-01       & 6.1019E-01       \\ \bottomrule
\end{tabular}}
\end{table}

%% file: Conclusion.tex
\section{Conclusion}
In this paper, we presented Chebyshev Feature Neural Network (CFNN), as a novel DNN structure for accurate function approximation.  CFNN  utilizes Chebyshev functions with learnable frequency in the first hidden layer. Combined with multi-stage learning and a proper initialization procedure, CFNN is capable of achieving machine accuracy for function approximation, which is a critical component for scientific machine learning.